\pdfoutput=1

\documentclass[11pt]{article}

\usepackage[preprint]{acl}
\usepackage{float}
\usepackage{arydshln}
\usepackage{makecell}
\usepackage{array}
\usepackage{graphicx} 
\usepackage{times}
\usepackage{latexsym}
\usepackage{booktabs}
\usepackage{multirow}
\newcommand\sr[1]{\textbf{\textcolor{purple}{SR: #1}}}

\usepackage{graphicx} 
\usepackage{tabularx}      
\usepackage{makecell}      
\usepackage{booktabs}      
\usepackage{caption} 
\usepackage{booktabs} 
\usepackage[T1]{fontenc}
\usepackage{afterpage}
\usepackage{amssymb} 
\usepackage[table]{xcolor}
\usepackage{arydshln} 

\usepackage[utf8]{inputenc}

\usepackage{microtype}

\usepackage{inconsolata}
\usepackage{float}

\usepackage{graphicx}

\usepackage{tcolorbox}

\usepackage[utf8]{inputenc}
\usepackage{tcolorbox}

\usepackage{amsmath}

\title{Leveraging Synthetic Data for Question Answering with Multilingual LLMs in the Agricultural Domain }

\author{First Author \\
  Affiliation / Address line 1 \\
  Affiliation / Address line 2 \\
  Affiliation / Address line 3 \\
  \texttt{email@domain} \\\And
  Second Author \\
  Affiliation / Address line 1 \\
  Affiliation / Address line 2 \\
  Affiliation / Address line 3 \\
  \texttt{email@domain} \\}

\author{
  \textbf{Rishemjit Kaur\textsuperscript{1,2,*}},
 \textbf{Arshdeep Singh Bhankhar\textsuperscript{1,*}},
  \textbf{Surangika Ranathunga\textsuperscript{3,*}},
  \textbf{Jashanpreet Singh Salh\textsuperscript{1}},
\\
  \textbf{Sudhir Rajput\textsuperscript{1}},
  \textbf{Vidhi\textsuperscript{1}},
  \textbf{Kashish Mahendra\textsuperscript{1}},
  \textbf{Bhavika Berwal\textsuperscript{1}},
  \textbf{Ritesh Kumar \textsuperscript{1,2}},
\\
 \textsuperscript{1}CSIR-Central Scientific Instruments Organisation, India
 \\
  \textsuperscript{2}Academy of Scientific and Innovative Research (AcSIR), Ghaziabad, U.P., India
  \\
  \textsuperscript{3}School of Mathematical and Computational Sciences, Massey University, New Zealand
  \\
  \small{
  \textsuperscript{*} These authors contributed equally}
\\
  \small{
    \textbf{Correspondence:} \href{mailto:rishemjit.kaur@csio.res.in,s.ranathunga@massey.ac.nz }{rishemjit.kaur@csio.res.in, s.ranathunga@massey.ac.nz}
  }
}

\begin{document}
\maketitle
\begin{abstract}
Enabling farmers to access accurate agriculture-related information in their native languages in a timely manner is crucial for the success of the agriculture field. Publicly available general-purpose  Large Language Models (LLMs) typically offer generic agriculture advisories, lacking precision in local and multilingual contexts. Our study addresses this limitation by generating multilingual (English, Hindi, Punjabi) synthetic  datasets from agriculture-specific documents from India and fine-tuning LLMs for the task of question answering (QA). Evaluation on human-created datasets demonstrates significant improvements in factuality, relevance, and agricultural consensus for the fine-tuned LLMs compared to the baseline counterparts. 

\end{abstract}

\section{Introduction}

Agriculture forms the backbone of many economies, specially for countries such as India. Despite substantial advancements of Artificial Intelligence across areas such as healthcare and education~\citep{wang2024large}, agriculture sector has seen relatively limited integration of cutting-edge technologies such as Large Language Models (LLMs). 
General-purpose LLMs have been shown to provide useful assistance in agricultural settings, but their advice is often generic, lacking in local specificity~\cite{ibrahim2023chatgpt,calone2024chatgpt}. Agriculture-specific chat-bot systems such as Farmer.Chat~\cite{singh2024farmer} that have multilingual support are promising, however there has been no quantitative evaluation of such systems. On the other hand, the handful of agriculture-specific Question Answer (QA) datasets are only for high resource languages such as English, German and Chinese~\cite{dofitas2025advanced,kasting2024assessing,jiang2025knowledge}.
In this paper, we present a novel synthetic QA dataset in three languages - English, Hindi and Punjabi, for the agriculture domain in India. With this synthetic training data, we carry out two different fine-tuning experiments (i) train and test LLMs with language-specific data (ii) train with English data and use translate-test approach~\cite{ahuja2023mega}, where the questions in Hindi/Punjabi are first translated to English, fed to the LLM trained with English data, and translating the generated responses back to Hindi/Punjabi. 

For evaluation, we provide human-curated datasets for all three languages. We carry out extensive human evaluations to determine the best LLM, as well as the prompt. Contrary to prior research that used either metrics such as ROGUE \cite{Johnny2022FarmerQuery, jiang2025knowledge} or LLM-as-judge \cite{yang2024gpt4, gauba2025agmmu} for agriculture QA evaluation, we employ human evaluation at each step.

Llama-3x instruct models emerged as the best for zero shot testing. Although they showed over 95\% for consensus, relevancy and factuality results were rather low, with the average across languages being 44.3\% and 29\%, respectively. However, after fine-tuning with synthetic datasets, these averages increased by 22.1\% and 14\% (respectively). The translate-test approach further increased results for Hindi and Punjabi.
\\Our contributions are: a synthetic training dataset for three languages, a human curated evaluation dataset, and a human evaluation on multiple LLMs and prompts, followed by insights on the inadequacy of automated evaluation schemes for domain- and country-specific QA evaluation tasks.

\section{Dataset Creation}
\subsection {Synthetic Training Data Generation}
The agricultural data used to generate QA pairs was collected through web scraping from Vikaspedia\footnote{\url{https://en.vikaspedia.in/}}, an online encyclopedia created by the Government of India. It contains data across diverse topics such as fisheries and women in agriculture. 
A summary of data source files and token statistics across these topics is presented in Table~\ref{tab:token-distribution} in Appendix~\ref{sec:data_source}. 

The raw text was split into  chunks of 2000 characters each, with an overlap of 200 characters between consecutive chunks. {These chunk and overlap sizes were selected after experimenting with multiple chunk
sizes (250–2000 chars) and overlap settings
(50–300 chars); smaller chunks degraded QA quality by limiting information inclusion, while larger overlaps showed no further gains.}

~\citet{paula-etal-2024-evaluation} showed that QA generation by prompting an LLM in  a few-shot setting is more effective than  approaches such as end-to-end generation (sequence of questions and answers is generated given the input context) and pipeline methods (answers are extracted first and questions are then generated in an answer-aware manner) and was therefore used for QA generation in our work.

{ We experimented with \textit{DeepSeek-R1-Distill-Llama-70B} \citep{guo2025deepseek} and \textit{Llama-3.1 405B} to create synthetic QA pairs. Former was selected due its better grammar and overall fluency. It provides more diverse question formulations and better ways to phrase questions and answers as compared to LLama 405 Billion Model} (Appendix~\ref{appendix:dSynthetic QA pair comparison between Llama and DeepSeek} shows sample QA pairs generated by each LLM). 

Figure~\ref{fig:prompt} in Appendix~\ref{appendix:prompt_synthethic_box} shows the prompt we designed to generate QA pairs from a given text segment of fixed chunk size using few-shot prompting. This prompt design was inspired by several past research~\citep{balaguer2024ragvsfinetuningpipelines, yehudai2024genie}. Contrary to previous research that instructed the LLM to generate a fixed number of questions~\cite{paula-etal-2024-evaluation}, we instruct it to generate as many questions as possible. This caters to the varying information density in the text chunks.

A total of 60,130 question-answer pairs were created from the agricultural dataset in English. This dataset was further translated into Hindi and Punjabi, using IndicTrans2~\cite{gala2023indictrans2}. \\

\subsection{Evaluation Dataset}

We manually created an English test dataset by collecting 1674 QA pairs from multiple sources such as Tamil Nadu Agricultural University, Union Bank of India-FAQ-Agri-Products, ICAR-NRRI, Cuttack, and ICAR-NRCP, Solapur. The links to sources are provided in Appendix \ref{appendix:dataCreators}.

{
A Hindi and Punjabi test set of 1219 QA pairs was developed by  manually transcribing 50 videos from the DD Kisan YouTube channel\footnote{\url{https://www.youtube.com/@DDKisan}}. The transcriptions were used to generate Hindi QA pairs, which were subsequently translated into Punjabi using IndicTrans2; QA generation methodology is elaborated in Appendix~\ref{appendix:transcription-methodology}. Hindi and Punjabi datasets were manually cleaned and verified to ensure an accurate and high quality test dataset.}

We validated the English-Hindi and Hindi-Punjabi  translation quality using human direct assessment~\cite{bojar2016findings}.The mean adequacy scores were 87.26 (En$\rightarrow$Hi) and 89.51 (Hi$\rightarrow$Pa). The inter-rater standard deviation was 5.72. Z-scores were computed for standardization, yielding mean values close to zero $(\approx 10^{-16})$.
 Further evaluation results and Translation error analysis are provided in Appendices~\ref{appendix : Translation and Transcription Quality for Hindi and Punjabi} and~\ref{appendix : Translation Error Analysis and Quality Control}, respectively.

\section{Implementation Details}
\subsection{Experiments}
\textbf{Selection of LLMs:} {LLMs differ in their language coverage. We therefore compared different LLMs to identify the best-performing one for multilingual agricultural QA. The LLMs selected  for initial zero-shot experiments are shown in Table~\ref{tab:models_by_language_trained}.{ Further details about LLMs are in Appendix~\ref{appendix:experimentation_details}}.}

\begin{table}[h]
\centering
\tiny
\begin{tabular}{lll}
\toprule
\textbf{Punjabi} & \textbf{Hindi} & \textbf{English} \\
\midrule
\textbf{Llama 3.1 8B}         & \textbf{Llama 3.1 8B}        & {{Llama 3.1 8B}} \\
Llama 3 8B           & Llama 3 8B          & \textbf{Llama 3 8B} \\
\textit{Navarasa 7B}          & \textit{Navarasa 7B}         & Deepseek 8B \\
  & Mistral 7B          & Mistral 7B \\ 
    & Gemma 8B            & {\textit{Airavata 7B}} \\
\bottomrule
\end{tabular}
\caption{LLMs selected for each language. The best performing LLM for each language is in \textbf{bold}. LLMs specifically trained for Indian languages are in \textit{italics}.}
\label{tab:models_by_language_trained}
\end{table}

\noindent\textbf{Prompt Selection:} 
We started with ~\citet{silva2023gpt4}'s prompt, which was originally designed for agriculture examination contexts. However, our long-term goal is to support farmers with advisories. Hence, we adapted that prompt and generated six different prompt variations, which are discussed in Appendix~\ref{appendix:prompts}.
Zero-shot experiments were conducted to select the best prompt.  

\begin{table*}
\centering
\tiny
\begin{tabular}{|c|p{2.5cm}|p{8cm}|p{3cm}|}
\hline
\textbf{No.} & \textbf{Parameter} & \textbf{Evaluation Question} & \textbf{Scale} \\
\hline
1 & Agricultural Consensus (Con) & How does the answer relate to the consensus of the agricultural domain? & Yes / No \\
\hline
2 & Relevancy (Rel) & Does the answer provide information pertinent to the question, regardless of whether the facts are entirely accurate? & 1 -- 5 \\
\hline
3 & Factuality (Fact) & Is the response generated correct, partially correct, or incorrect? Does it contain any content it shouldn’t? & Correct / Incorrect / Partially Correct \\
\hline
\end{tabular}
\caption{Evaluation criteria for human assessment}
\label{tab:evaluation_criteria}
\end{table*}

\textbf{LLM Fine-tuning:} The best performing LLM  under the zero-shot setting was fine-tuned with the best prompt, using the synthetic dataset we created.  We also conducted an experiment to evaluate the efficacy of the translate-test approach for Hindi and Punjabi. Implementation details and hyperparameters are in Appendix~\ref{appendix:hyperparameters}.

\subsection{Evaluation Procedure}
From each QA pair in the test set, the question was used in the prompt, and the corresponding answer generated by the LLM was evaluated against the ground-truth answer by a human with respect to three metrics (see Table~\ref{tab:evaluation_criteria}): Agricultural Consensus, Relevancy, and Factuality. These metrics are inspired by similar evaluation methods in prior research~\cite{singhal2023large,balaguer2024ragvsfinetuningpipelines}. 

Due to the scarcity of human evaluators, it was not possible to evaluate all the LLMs across the full test set. Therefore, a subset of the QA pairs (see Table~\ref{tab:experimentationx}) was evaluated by three annotators. Multiple raters/measurements~\cite{KOO2016155} was used to calculate inter-annotator agreement for relevancy and factuality. Fleiss Kappa ~\cite{fleiss1971kappa} was used for agriculture consensus. Human evaluator details and evaluation guidelines are outlined in Appendix ~\ref{appendix:HME}. Why these measures were selected and how to interpret the results  are  in Appendix~\ref{sec:iaa}. Inter-rater agreement is reported in Table~\ref{tab:inter_rater_agreement}. Agreement for consensus is almost perfect, while agreement for relevancy and factuality show moderate to good agreement.

\begin{table}[t]
\centering
\normalsize
\resizebox{\linewidth}{!}{%
\begin{tabular}{lcccc}
\toprule
\textbf{Setting} & \textbf{LLMs/} & \textbf{Temp} & \textbf{Unique} & \textbf{Total} \\
 & \textbf{Prompts} &  & \textbf{QA pairs} & \textbf{Evals} \\
\midrule
\multicolumn{5}{l}{\textit{Multilingual LLM Comparison}} \\
English & 4 LLMs & 3 & 20 & 720\textsuperscript{*}\\
Hindi & 5 LLMs & 1 & 100 & 1500\textsuperscript{*}\\
Punjabi & 3 LLMs & 1 & 100 & 900\textsuperscript{*}\\

\midrule
\multicolumn{5}{l}{\textit{Prompt Selection (1 LLM)}} \\
English & 6 Prompts & 1 & 20 & 120 \\
\midrule
\multicolumn{5}{l}{\textit{Base vs. Finetuned Evaluation}} \\
English & 2 (Base+FT-EN) & 1 & 300 & 600 \\
Hindi & 2 (Base+FT-HI) & 1 & 240 & 480 \\
Punjabi & 2 (Base+FT-PA) & 1 & 240 & 480 \\
\midrule
\multicolumn{5}{l}{\textit{Translate Test Approach (via FT-EN model)}} \\
Hindi & 1 (FT-EN) & 1 & 240 & 240 \\
Punjabi & 1 (FT-EN) & 1 & 240 & 240 \\
\midrule
\textbf{Total Evaluations} & & & & \textbf{5,300} \\
\bottomrule
\end{tabular}
}
\caption{Number of QA samples used for human experiments.  Temperature values 0.2, 0.5, 0.7 were tested only for English model selection.*Each QA pair was independently evaluated by 3 annotators. }
\label{tab:experimentationx}
\end{table}

\section{Results}

\subsection{Model and Prompt Selection }
Among the six tested prompts, Prompt 1 (see appendix~\ref{appendix:prompt_qa_box}) achieved the highest scores for English with 100\%, 95.0\% and 55.0\% in consensus, relevancy and factuality (respectively) with Llama-3 model and was therefore used in the subsequent experiments. Results for all the prompts are in Table~\ref{tab:human_eval_results_prompt} of Appendix~\ref{appendix:ModelTempSelection}.

Results in Table \ref{tab:model_selection_result} show that \textit{Llama-3-8B-Instruct}, \textit{Llama-3.1-8B-Instruct} and \textit{Llama-3.1-8B-Instruct} emerged as the best LLMs for English, Hindi and Punjabi, respectively, at temperature T = 0.2. Experiments were also conducted at T = 0, 0.5 and 0.7 for English (for results see Table \ref{tab:temp_comparison} and \ref{tab:model-eval-temp} of Appendix~\ref{appendix:ModelTempSelection}). Temperature was varied as models exhibit different optimal generation settings. For instance, \citet{deepseek_temperature_param} recommends 0.0 for coding/math but higher values ($\approx (1.3 \text{--}1.5)$) for conversation or creative tasks, and some models are also sensitive to prompt style or domain.

{ It was observed that for English, \textit{Llama-3} produced the most accurate and relevant responses and was free from repetition or gibberish content. \textit{Airavata} often produced overly brief answers with repeated points, \textit{Mistral} recalled information accurately but hallucinated incorrect facts, and \textit{DeepSeek} responses were unnecessarily long. For Hindi and Punjabi, some models returned incomplete answers and code-mixed text. Punjabi responses also occasionally contained Shahmukhi (Urdu script) instead of Gurmukhi for certain words. While these issues were also observed in \textit{Llama-3.1}, they occurred significantly less frequently compared to other models.}

\begin{table}[t]
\tiny
\setlength{\tabcolsep}{4pt} 
\centering

\begin{tabularx}{\columnwidth}{@{}lXccc@{}}
\toprule
\textbf{Metric} & \textbf{Calculation} & \textbf{English} & \textbf{Hindi} & \textbf{Punjabi} \\
\midrule
\makecell[l]{Consensus} & Fleiss' $\kappa$ & 1.0 & 1.0 & 0.95 \\
Relevancy             & ICC   & 0.79 & 0.62 & 0.77 \\
Factuality            & ICC   & 0.82 & 0.93 & 0.75 \\
\bottomrule
\end{tabularx}
\caption{Inter-rater agreement for the three languages.}
\label{tab:inter_rater_agreement}
\end{table}

\begin{table}[t]
\centering
\tiny
\begin{tabular}{|c|c|c|c|c|}
\hline
\textbf{Lang.} & \textbf{LLM} & \textbf{Con.(\%)} & \textbf{Rel.\%)} & \textbf{Fact.(\%)} \\
\hline
\multirow{4}{*}{\rotatebox[origin=c]{90}{\textbf{English}}} & Airavata-7B-Instruct & 100.0 & 58.3 & 30.0 \\
& DeepSeek-8B-Instruct& 100.0 & 33.3 & 28.3 \\
& \textbf{Llama-3-8B-Instruct} & \textbf{100.0} & \textbf{98.3} & \textbf{66.6} \\
& \text{Llama-3.1-8B-Instruct} & \text{95.0} & \text{58.3} & \text{16.6}\\
& Mistral-7B-Instruct & 98.3 & 93.3 & 30.0 \\
\hline

\multirow{5}{*}{\rotatebox[origin=c]{90.0}{\textbf{Hindi}}} & Gemma-7B-Instruct & 90 & 26.3 & 14.6 \\
& Llama-3-8B-Instruct & 84.0 & 25.3 & 13.3 \\
& \textbf{Llama-3.1-8B-Instruct} & \textbf{96.0} & \textbf{74.6} & \textbf{35.6}\\
& Mistral-7B-Instruct & 100.0 & 37.3 & 5.6 \\
& Navarasa-7B-Instruct & 91.0 & 58.6 & 21.0 \\
\hline

\multirow{3}{*}{\rotatebox[origin=c]{90}{\textbf{Punjabi}}} & Llama-3-8B-Instruct & 87.0 & 17.6& 12.3 \\
& \textbf{Llama-3.1-8B-Instruct} & \textbf{100.0} & \textbf{43.3} & \textbf{24.3} \\
& Navarasa-7B-Instruct & 95.6& 29.3 & 15.6\\
\hline
\end{tabular}
\caption{Human evaluation results for model selection across languages at temperature \(T = 0.2\)}
\label{tab:model_selection_result}
\end{table}

\subsection{Base vs. Fine-Tuned LLM Results}

For each language, the LLM that provided the best result in the zero-shot setup was used for fine-tuning experiments. Results are in Table~\ref{tab:base_vs_finetuned_results}. A notable improvement was observed for relevancy and factuality across all languages after fine-tuning, despite a drop in performance for consensus for Hindi and Punjabi.
The results also indicate that the translate-test approach outperformed the Hindi and Punjabi fine-tuned LLMs. Results are the best for English, followed by Hindi and Punjabi. {This is because LLMs are heavily trained on English data, and the translate-test approach allows the model to operate in its strongest language setting. This could be a result of differing data amounts used in LLM pre-training as stated by \citet{ahuja2023mega}.}

\begin{table}[t]
\centering
\tiny
\setlength{\tabcolsep}{5pt} 
\renewcommand{\arraystretch}{1.4} 

\begin{tabular}{|c|ccc|ccc|ccc|}
\hline
\textbf{Lang} & \multicolumn{3}{c|}{\textbf{Consensus(\%)}} 
& \multicolumn{3}{c|}{\textbf{Relevancy(\%)}} 
& \multicolumn{3}{c|}{\textbf{Factuality(\%)}} \\
& \textbf{BASE} & \textbf{FT} & \textbf{TT} 
  & \textbf{BASE} & \textbf{FT} & \textbf{TT} 
  & \textbf{BASE} & \textbf{FT} & \textbf{TT} \\
\hline
En &  98.0 & \textbf{99.3} & --    & 32.3 & \textbf{70.6} & --    & 32.6& \textbf{55.3} & -- \\
Hi & \textbf{97.5} & 93.3 & 96.2 & 62.5 & 66.6 & \textbf{74.1} & 28.7 & 38.3 & \textbf{42.0} \\
Pa & {95.0} & 85.4 & \textbf{96.6} & 38.3 & 62.5 & \textbf{72.9} & 25.8 & 35.4 & \textbf{37.9} \\
\hline
\end{tabular}

\caption{Comparison of Base, Fine-Tuned (FT) and Translate-test with FT-EN LLM (TT) performance.}
\label{tab:base_vs_finetuned_results}
\end{table}

\section{Challenges with LLM-as-a-Judge Evaluation}
In the above experiments, 5,280 human evaluations were done. Although human evaluations are the gold standard for assessing LLM performance, they are time consuming and expensive~\cite{tan2024judgebench}. As a solution, we employed  the "LLM-as-a-Judge" approach for evaluation.
We used three instruction-tuned LLMs, namely \textit{Llama-4-Scout-17B-16E-Instruct},\textit{ Llama-4-Maverick-17B-128E-Instruct}, and \textit{DeepSeek-R1-Distill-Llama-70B}\footnote{We excluded larger LLMs like \textit{Llama 4 Behemoth} (288B) and \textit{DeepSeek-R1} (685B) due to resource constraints.}.
Then we generated separate prompts (See Appendix ~\ref{appendix:llmjudge-prompts}), each corresponding to an evaluation metric, which were adapted from G-eval ~\citep{liu2023geval}.  The QA pairs used for human evaluation were evaluated using LLM judges.

 As illustrated by examples provided in Table~\ref{tab:qa-eval-full} of Appendix~\ref{appendix:misleadingSCORES}, the LLM judge misclassifies LLM response as partially correct. Bertscore, ROUGE and BLEU values are inconsistent, making it difficult to reach a judgment. Correlation between LLM judges and humans is shown in Table~\ref{tab:llm-human-correlation}. Humans and the LLM Judges had a near-perfect agreement for consensus (see Appendix~\ref{appendix:PearsonAgriConsensus}). However, even the best result produced by the LLM judges showed a weak or moderate correlation for both relevance and factuality. {The factuality metric was assessed on a 3-point categorical scale whereas the relevance metric was a Likert scale (1-5). As shown in previous research \cite{shen-etal-2023-large, liu2023geval}, LLM judges often tend to concentrate on a single rating on wide scales, leading to low variance and weak correlation with human judgment. In our case, the majority of LLM judgments were concentrated at the highest Likert point (5) and resulted in low correlation for relevance (0.18) as compared to factuality (0.46),
 questioning the validity of LLM-as-a-Judge for domain, language and country-specific QA tasks. }

\begin{table}[H]
\centering
\resizebox{\linewidth}{!}{%
\begin{tabular}{llccc}
\toprule
\textbf{Lang.} & \textbf{LLM} & \textbf{Rel. (Spearman $\rho$)} & \textbf{Fact. (Spearman $\rho$)} \\
\midrule
\multirow{3}{*}{En} 
 & DeepSeek     & 0.23 & 0.27 \\
 & L4-Scout     & 0.18 & \textbf{0.46} \\
 & L4-Maverick  & \textbf{0.24} & 0.37 \\
\midrule
\multirow{3}{*}{Hi} 
 & DeepSeek     & \textbf{0.49} & 0.44 \\
 & L4-Scout     & 0.45 & \textbf{0.49} \\
 & L4-Maverick  & \textbf{0.49} & 0.45 \\
\midrule
\multirow{3}{*}{Pa} 
 & DeepSeek     & 0.35 & 0.23 \\
 & L4-Scout     & \textbf{0.40} & \textbf{0.39} \\
 & L4-Maverick  & 0.39 & 0.38 \\
\bottomrule
\end{tabular}%
}
\caption{Correlation between LLMs and humans.} 
\label{tab:llm-human-correlation}
\end{table}

\section{Conclusion}

We introduced a synthetic fine-tuning dataset and a human-curated evaluation dataset in English, Hindi, and Punjabi for QA in the Indian agricultural context. Our experiments demonstrate that fine-tuned LLMs significantly outperform their base counterparts, underscoring the value of synthetic data. Furthermore, the translate-test approach yields better performance than directly fine-tuning LLMs in Hindi or Punjabi. We hope that the insights on the inadequacy of LLM judges for domain, language
and country-specific QA tasks will caution future researchers and promote further research to overcome the associated issues.  

\section{Limitations}
There are several limitations in our  work. First, the dataset created and the methodology used in this study include only textual data. However, in real-world scenarios, farmers may have image-based queries such as pictures of crops or pests/diseases. It is not addressed by our current approach. Our future work will focus on building a multimodal and multilingual dataset that can handle both textual and visual input. It will also incorporate a wider range of Indian languages. Second, human evaluation was done only on a subset of the dataset, due to the resource-intensive nature of the task. In future, a more comprehensive human evaluation across the entire dataset is  necessary to ensure LLM's efficacy and deployment. Another limitation is that our experiments were conducted using relatively smaller LLMs (7B to 8B parameter range) due to limited computational resources. In future, the effect of using larger LLMs on performance can be explored.  Our current work does not account for dynamic contextual factors such as weather conditions or soil health, which are often important in generating accurate and dynamic advisories. Integrating such contextual data through various APIs in an agentic framework can enable adaptive and real-time responses and it is an important direction for our future work. As the next step, we plan to further validate our results through field testing with farmers. We also plan to explore techniques such as RAG or LLM agents to get up-to-date information on current weather, crop prices, etc.

\section{Ethical Considerations}
The human annotators volunteered their time to contribute to the project, and no financial compensation was provided. Our approach follows the principles of participatory research~\cite{nekoto2020participatory, gamage2025multilingual}, and data creators have been offered coauthorship to compensate for their contributions. We crawled the data by adhering to the robots.txt specifications for each website.

{This study uses data obtained from publicly accessible Government of India websites, including Vikaspedia and official ICAR portals, all of which are properly cited. The intellectual property rights of the content remain with the respective original sources. The use of the data complies with the fair dealing provisions of the Indian Copyright Act, which permit the use of copyrighted material for non-commercial research purposes. To ensure compliance , we have also sought approvals from all the sources. We have already received approval for Vikaspedia, which was used for creating fine tuning data and can be freely re-distributed. The evaluation datasets will be released publicly upon receiving the necessary approvals from the respective data providers.}

\bibliography{custom}

@article{gauba2025agmmu,
  title={AgMMU: A Comprehensive Agricultural Multimodal Understanding and Reasoning Benchmark},
  author={Gauba, Aruna and Pi, Irene and Man, Yunze and Pang, Ziqi and Adve, Vikram S and Wang, Yu-Xiong},
  journal={arXiv preprint arXiv:2504.10568},
  year={2025}
}

@article{wang2024large,
  title={Large language models for education: A survey and outlook},
  author={Wang, Shen and Xu, Tianlong and Li, Hang and Zhang, Chaoli and Liang, Joleen and Tang, Jiliang and Yu, Philip S and Wen, Qingsong},
  journal={arXiv preprint arXiv:2403.18105},
  year={2024}
}

@misc{balaguer2024ragvsfinetuningpipelines,
      title={RAG vs Fine-tuning: Pipelines, Tradeoffs, and a Case Study on Agriculture}, 
      author={Angels Balaguer and Vinamra Benara and Renato Luiz de Freitas Cunha and Roberto de M. Estevão Filho and Todd Hendry and Daniel Holstein and Jennifer Marsman and Nick Mecklenburg and Sara Malvar and Leonardo O. Nunes and Rafael Padilha and Morris Sharp and Bruno Silva and Swati Sharma and Vijay Aski and Ranveer Chandra},
      year={2024},
      eprint={2401.08406},
      archivePrefix={arXiv},
      primaryClass={cs.CL},
      url={https://arxiv.org/abs/2401.08406}, 
}

@article{yehudai2024genie,
  title={Genie: Achieving human parity in content-grounded datasets generation},
  author={Yehudai, Asaf and Carmeli, Boaz and Mass, Yosi and Arviv, Ofir and Mills, Nathaniel and Toledo, Assaf and Shnarch, Eyal and Choshen, Leshem},
  journal={arXiv preprint arXiv:2401.14367},
  year={2024}
}

@article{guo2025deepseek,
  title={Deepseek-r1: Incentivizing reasoning capability in llms via reinforcement learning},
  author={Guo, Daya and Yang, Dejian and Zhang, Haowei and Song, Junxiao and Zhang, Ruoyu and Xu, Runxin and Zhu, Qihao and Ma, Shirong and Wang, Peiyi and Bi, Xiao and others},
  journal={arXiv preprint arXiv:2501.12948},
  year={2025},
  url          = {https://arxiv.org/abs/2501.12948},
}

@inproceedings{paula-etal-2024-evaluation,
    title = "Evaluation of Question Answer Generation for {P}ortuguese: Insights and Datasets",
    author = "Paula, Felipe  and
      Michelin, Cassiana Roberta Lizzoni  and
      Moreira, Viviane",
    editor = "Al-Onaizan, Yaser  and
      Bansal, Mohit  and
      Chen, Yun-Nung",
    booktitle = "Findings of the Association for Computational Linguistics: EMNLP 2024",
    year = "2024",
    address = "Miami, Florida, USA",
    publisher = "Association for Computational Linguistics",
    url = "https://aclanthology.org/2024.findings-emnlp.306/",
    doi = "10.18653/v1/2024.findings-emnlp.306",
    pages = "5315--5327",
}

@article{singhal2023large,
  title={Large language models encode clinical knowledge},
  author={Singhal, Karan and Azizi, Shekoofeh and Tu, Tao and Mahdavi, S Sara and Wei, Jason and Chung, Hyung Won and Scales, Nathan and Tanwani, Ajay and Cole-Lewis, Heather and Pfohl, Stephen and others},
  journal={Nature},
  volume={620},
  number={7972},
  pages={172--180},
  year={2023},
  publisher={Nature Publishing Group},
  url = "https://doi.org/10.1038/s41586-023-06291-2",
  doi = "10.1038/s41586-023-06291-2"
}

@article{ahuja2023mega,
  title={Mega: Multilingual evaluation of generative ai},
  author={Ahuja, Kabir and Diddee, Harshita and Hada, Rishav and Ochieng, Millicent and Ramesh, Krithika and Jain, Prachi and Nambi, Akshay and Ganu, Tanuja and Segal, Sameer and Axmed, Maxamed and others},
  journal={arXiv preprint arXiv:2303.12528},
  year={2023}
}

@misc{deepseek_temperature_param,
  author       = {{DeepSeek, Inc.}},
  title        = {The Temperature Parameter},
  year         = {2025},
  url          = {https://api-docs.deepseek.com/quick_start/parameter_settings},
  note   = {Accessed: January 1, 2025},
}

@inproceedings{bojar2016findings,
    title = "Findings of the 2016 Conference on Machine Translation",
    author = "Bojar, Ond{\v{r}}ej  and
      Chatterjee, Rajen  and
      Federmann, Christian  and
      Graham, Yvette  and
      Haddow, Barry  and
      Huck, Matthias  and
      Jimeno Yepes, Antonio  and
      Koehn, Philipp  and
      Logacheva, Varvara  and
      Monz, Christof  and
      Negri, Matteo  and
      N{\'e}v{\'e}ol, Aur{\'e}lie  and
      Neves, Mariana  and
      Popel, Martin  and
      Post, Matt  and
      Rubino, Raphael  and
      Scarton, Carolina  and
      Specia, Lucia  and
      Turchi, Marco  and
      Verspoor, Karin  and
      Zampieri, Marcos",
    editor = {Bojar, Ond{\v{r}}ej  and
      Buck, Christian  and
      Chatterjee, Rajen  and
      Federmann, Christian  and
      Guillou, Liane  and
      Haddow, Barry  and
      Huck, Matthias  and
      Yepes, Antonio Jimeno  and
      N{\'e}v{\'e}ol, Aur{\'e}lie  and
      Neves, Mariana  and
      Pecina, Pavel  and
      Popel, Martin  and
      Koehn, Philipp  and
      Monz, Christof  and
      Negri, Matteo  and
      Post, Matt  and
      Specia, Lucia  and
      Verspoor, Karin  and
      Tiedemann, J{\"o}rg  and
      Turchi, Marco},
    booktitle = "Proceedings of the First Conference on Machine Translation: Volume 2, Shared Task Papers",
    month = aug,
    year = "2016",
    address = "Berlin, Germany",
    publisher = "Association for Computational Linguistics",
    url = "https://aclanthology.org/W16-2301/",
    doi = "10.18653/v1/W16-2301",
    pages = "131--198"
}

@inproceedings{rohera2024indicquest,
    title = "{L}3{C}ube-{I}ndic{Q}uest: A Benchmark Question Answering Dataset for Evaluating Knowledge of {LLM}s in {I}ndic Context",
    author = "Rohera, Pritika  and
      Ginimav, Chaitrali  and
      Salunke, Akanksha  and
      Sawant, Gayatri  and
      Joshi, Raviraj",
    editor = "Oco, Nathaniel  and
      Dita, Shirley N.  and
      Borlongan, Ariane Macalinga  and
      Kim, Jong-Bok",
    booktitle = "Proceedings of the 38th Pacific Asia Conference on Language, Information and Computation",
    month = dec,
    year = "2024",
    address = "Tokyo, Japan",
    publisher = "Tokyo University of Foreign Studies",
    url = "https://aclanthology.org/2024.paclic-1.93/",
    pages = "982--988"
}

@article{ALONSO2024102938,
title = {MedExpQA: Multilingual benchmarking of Large Language Models for Medical Question Answering},
journal = {Artificial Intelligence in Medicine},
volume = {155},
pages = {102938},
year = {2024},
issn = {0933-3657},
doi = {https://doi.org/10.1016/j.artmed.2024.102938},
url = {https://www.sciencedirect.com/science/article/pii/S0933365724001805},
author = {Iñigo Alonso and Maite Oronoz and Rodrigo Agerri}
}

@inproceedings{shen-etal-2023-large,
    title     = {Large Language Models Are Not Yet Human-Level Evaluators for Abstractive Summarization},
    author    = {Shen, Chen and Cheng, Lian and Nguyen, Xuan-Phi and You, Yang and Bing, Lidong},
    booktitle = {Findings of the Association for Computational Linguistics: EMNLP 2023},
    month     = dec,
    year      = {2023},
    address   = {Singapore},
    publisher = {Association for Computational Linguistics},
    pages     = {4215--4233},
    doi       = {10.18653/v1/2023.findings-emnlp.278},
    url       = {https://aclanthology.org/2023.findings-emnlp.278}
}

@ARTICLE{dofitas2025advanced,
  author={Dofitas, Cyreneo and Kim, Yong-Woon and Byun, Yung-Cheol},
  journal={IEEE Access}, 
  title={Advanced Agricultural Query Resolution Using Ensemble-Based Large Language Models}, 
  year={2025},
  volume={13},
  number={},
  pages={34732-34746},
  doi={10.1109/ACCESS.2025.3541602}}

@inproceedings{nekoto2020participatory,
    title = "Participatory Research for Low-resourced Machine Translation: A Case Study in {A}frican Languages",
    author = {Nekoto, Wilhelmina  and
      Marivate, Vukosi  and
      Matsila, Tshinondiwa  and
      Fasubaa, Timi  and
      Fagbohungbe, Taiwo  and
      Akinola, Solomon Oluwole  and
      Muhammad, Shamsuddeen  and
      Kabongo Kabenamualu, Salomon  and
      Osei, Salomey  and
      Sackey, Freshia  and
      Niyongabo, Rubungo Andre  and
      Macharm, Ricky  and
      Ogayo, Perez  and
      Ahia, Orevaoghene  and
      Berhe, Musie Meressa  and
      Adeyemi, Mofetoluwa  and
      Mokgesi-Selinga, Masabata  and
      Okegbemi, Lawrence  and
      Martinus, Laura  and
      Tajudeen, Kolawole  and
      Degila, Kevin  and
      Ogueji, Kelechi  and
      Siminyu, Kathleen  and
      Kreutzer, Julia  and
      Webster, Jason  and
      Ali, Jamiil Toure  and
      Abbott, Jade  and
      Orife, Iroro  and
      Ezeani, Ignatius  and
      Dangana, Idris Abdulkadir  and
      Kamper, Herman  and
      Elsahar, Hady  and
      Duru, Goodness  and
      Kioko, Ghollah  and
      Espoir, Murhabazi  and
      van Biljon, Elan  and
      Whitenack, Daniel  and
      Onyefuluchi, Christopher  and
      Emezue, Chris Chinenye  and
      Dossou, Bonaventure F. P.  and
      Sibanda, Blessing  and
      Bassey, Blessing  and
      Olabiyi, Ayodele  and
      Ramkilowan, Arshath  and
      {\"O}ktem, Alp  and
      Akinfaderin, Adewale  and
      Bashir, Abdallah},
    editor = "Cohn, Trevor  and
      He, Yulan  and
      Liu, Yang",
    booktitle = "Findings of the Association for Computational Linguistics: EMNLP 2020",
    month = nov,
    year = "2020",
    address = "Online",
    publisher = "Association for Computational Linguistics",
    url = "https://aclanthology.org/2020.findings-emnlp.195/",
    doi = "10.18653/v1/2020.findings-emnlp.195",
    pages = "2144--2160",
    
}

@ARTICLE{gamage2025multilingual,
  author={Gamage, Omega and Ranathunga, Surangika and Lee, Annie and Sun, Xiao and Singh, Aryaveer and Skenduli, Marjana Prifti and Alam, Mehreen and Nayak, Ajit Kumar and Gao, Haonan and Deori, Barga and Ji, Jingwen and Zhang, Qiyue and Zeng, Yuchen and Tian, Muxin and Mao, Yanke and Trico, Endi and Nako, Danja and Shqezi, Sonila and Hoxha, Sara and Imami, Dezi and Doksani, Dea and Pandey, Virat Kumar and Ananya, Ananya and Aggarwal, Nitisha and Hussain, Naiyarah and Dwivedi, Vandana and Sinha, Rajkumari Monimala and Kalita, Dhrubajyoti},
  journal={IEEE Transactions on Audio, Speech and Language Processing}, 
  title={A Multilingual Dataset (MultiMWP) and Benchmark for Math Word Problem Generation}, 
  year={2025},
  volume={33},
  number={},
  pages={1838-1848},
  doi={10.1109/TASLPRO.2025.3552936}}

@article{jiang2025knowledge,
title = {Knowledge assimilation: Implementing knowledge-guided agricultural large language model},
journal = {Knowledge-Based Systems},
volume = {314},
pages = {113197},
year = {2025},
issn = {0950-7051},
doi = {https://doi.org/10.1016/j.knosys.2025.113197},
url = {https://www.sciencedirect.com/science/article/pii/S0950705125002448},
author = {Jingchi Jiang and Lian Yan and Haifeng Liu and Zhenbo Xia and Haotian Wang and Yang Yang and Yi Guan},
}

@incollection{kasting2024assessing,
author = "Kästing, Marvin and Hänig, Christian",
title = "Assessing Large Language Models in the Agricultural Sector: A Comprehensive Analysis Utilizing a Novel Synthetic Benchmark Dataset",
year = 2024,
doi = "10.18420/inf2024_113",
booktitle = "INFORMATIK 2024",
publisher = "Gesellschaft für Informatik e.V.",
address = "Bonn",
issn = "2944-7682",
pissn = "1617-5468",
eissn = "2944-7682",
isbn = "978-3-88579-746-3",
pages = "1279--1286",
}

@inproceedings{Dettmers2023QLoRA,
author = {Dettmers, Tim and Pagnoni, Artidoro and Holtzman, Ari and Zettlemoyer, Luke},
title = {QLORA: efficient finetuning of quantized LLMs},
year = {2023},
publisher = {Curran Associates Inc.},
address = {Red Hook, NY, USA},
booktitle = {Proceedings of the 37th International Conference on Neural Information Processing Systems},
articleno = {441},
numpages = {28},
location = {New Orleans, LA, USA},
series = {NIPS '23}
}

@article{calone2024chatgpt,
title = {Analysing the potential of ChatGPT to support plant disease risk forecasting systems},
journal = {Smart Agricultural Technology},
volume = {10},
pages = {100824},
year = {2025},
issn = {2772-3755},
doi = {https://doi.org/10.1016/j.atech.2025.100824},
url = {https://www.sciencedirect.com/science/article/pii/S2772375525000589},
author = {Roberta Calone and Elisabetta Raparelli and Sofia Bajocco and Eugenio Rossi and Lorenzo Crecco and Danilo Morelli and Chiara Bassi and Rocchina Tiso and Riccardo Bugiani and Fabio Pietrangeli and Giovanna Cattaneo and Camilla Nigro and Marco Secondo Gerardi and Simone Bussotti and Angela Sanchioni and Danilo Tognetti and Mariangela Sandra and Irene {De Lillo} and Paolo Framarin and Sandra Di Ferdinando and Simone Bregaglio},
}

@article{ibrahim2023chatgpt,
  title={Evaluating responses by ChatGPT to farmers’ questions on irrigated lowland rice cultivation in Nigeria},
  author={Ibrahim, Ali and Senthilkumar, Kalimuthu and Saito, Kazuki},
  journal={Scientific Reports},
  volume={14},
  number={1},
  pages={3407},
  year={2024},
  URL  = { https://doi.org/10.1038/s41598-024-53916-1 },
DOI = { 10.1038/s41598-024-53916-1 },
  publisher={Nature Publishing Group UK London}
}

@article{
gala2023indictrans2,
title={IndicTrans2: Towards High-Quality and Accessible Machine Translation Models for all 22 Scheduled Indian Languages},
author={Jay Gala and Pranjal A Chitale and A K Raghavan and Varun Gumma and Sumanth Doddapaneni and Aswanth Kumar M and Janki Atul Nawale and Anupama Sujatha and Ratish Puduppully and Vivek Raghavan and Pratyush Kumar and Mitesh M Khapra and Raj Dabre and Anoop Kunchukuttan},
journal={Transactions on Machine Learning Research},
issn={2835-8856},
year={2023},
url={https://openreview.net/forum?id=vfT4YuzAYA},
note={}
}

@misc{google2025gemini2flash,
  author       = {{Google DeepMind}},
  title        = {Gemini 2.0 Flash},
  year         = {2025},
  howpublished = {\url{https://ai.google.dev/gemini-api/docs/models/gemini-2.0-flash}}
}

@article{silva2023gpt4,
  title        = {GPT‑4 as an Agronomist Assistant? Answering Agriculture Exams Using Large Language Models},
  author       = {Silva, Bruno and Nunes, Leonardo and Estev{\~a}o, Roberto and Aski, Vijay and Chandra, Ranveer},
  journal      = {arXiv preprint arXiv:2310.06225},
  year         = {2023},
}

@inproceedings{liu2023geval,
    title = "{G}-Eval: {NLG} Evaluation using Gpt-4 with Better Human Alignment",
    author = "Liu, Yang  and
      Iter, Dan  and
      Xu, Yichong  and
      Wang, Shuohang  and
      Xu, Ruochen  and
      Zhu, Chenguang",
    editor = "Bouamor, Houda  and
      Pino, Juan  and
      Bali, Kalika",
    booktitle = "Proceedings of the 2023 Conference on Empirical Methods in Natural Language Processing",
    month = dec,
    year = "2023",
    address = "Singapore",
    publisher = "Association for Computational Linguistics",
    url = "https://aclanthology.org/2023.emnlp-main.153/",
    doi = "10.18653/v1/2023.emnlp-main.153",
    pages = "2511--2522",
    
}

@article{KOO2016155,
title = {A Guideline of Selecting and Reporting Intraclass Correlation Coefficients for Reliability Research},
journal = {Journal of Chiropractic Medicine},
volume = {15},
number = {2},
pages = {155-163},
year = {2016},
issn = {1556-3707},
doi = {https://doi.org/10.1016/j.jcm.2016.02.012},
url = {https://www.sciencedirect.com/science/article/pii/S1556370716000158},
author = {Terry K. Koo and Mae Y. Li}
}

@article{fleiss1971kappa,
  title={Measuring nominal scale agreement among many raters.},
  author={Fleiss, Joseph L},
  journal={Psychological bulletin},
  volume={76},
  number={5},
  pages={378},
  year={1971},
  doi = {10.1037/h0031619},
  publisher={American Psychological Association}
}

@article{pearson1895correlation,
    author = {Pearson, Karl},
    title = {X. Contributions to the mathematical theory of evolution.—II. Skew variation in homogeneous material},
    journal = {Philosophical Transactions of the Royal Society of London, Series A: Containing Papers of a Mathematical or Physical Character},
    volume = {186},
    pages = {343-414},
    year = {1895},
    month = {12},
    issn = {0264-3952},
    doi = {10.1098/rsta.1895.0010},
    url = {https://doi.org/10.1098/rsta.1895.0010},
    eprint = {https://royalsocietypublishing.org/rsta/article-pdf/doi/10.1098/rsta.1895.0010/1272717/rsta.1895.0010.pdf},
}

@inproceedings{
tan2024judgebench,
title={JudgeBench: A Benchmark for Evaluating {LLM}-Based Judges},
author={Sijun Tan and Siyuan Zhuang and Kyle Montgomery and William Yuan Tang and Alejandro Cuadron and Chenguang Wang and Raluca Popa and Ion Stoica},
booktitle={The Thirteenth International Conference on Learning Representations},
year={2025},
url={https://openreview.net/forum?id=G0dksFayVq}
}

@inproceedings{fu-etal-2023-large,
  title     = "Are Large Language Models Reliable Judges? A Study on the Factuality Evaluation Capabilities of {LLM}s",
  author    = "Fu, Xue-Yong and Laskar, Md Tahmid Rahman and Chen, Cheng and Tn, Shashi Bhushan",
  editor    = "Gehrmann, Sebastian and Wang, Alex and Sedoc, Jo{\~a}o and Clark, Elizabeth and Dhole, Kaustubh and Chandu, Khyathi Raghavi and Santus, Enrico and Sedghamiz, Hooman",
  booktitle = "Proceedings of the Third Workshop on Natural Language Generation, Evaluation, and Metrics (GEM)",
  month     = dec,
  year      = "2023",
  address   = "Singapore",
  publisher = "Association for Computational Linguistics",
  pages     = "310--316",
  url       = "https://aclanthology.org/2023.gem-1.25/",
}

@article{landis-koch-1977-measurement,
  author       = {Landis, J.\ Richard and Koch, G.\ Gary},
  title        = {The Measurement of Observer Agreement for Categorical Data},
  journal      = {Biometrics},
  volume       = {33},
  number       = {1},
  pages        = {159--174},
  year         = {1977},
  publisher    = {International Biometric Society},
  doi          = {10.2307/2529310},
  url          = {https://www.jstor.org/stable/2529310}
}

@article{singh2024farmer,
author = {Singh, Namita and Wang'ombe, Jacqueline and Okanga, Nereah and Zelenska, Tetyana and Repishti, Jona and K, Jayasankar G and Mishra, Sanjeev and Manokaran, Rajsekar and Singh, Vineet and Rafiq, Mohammed Irfan and Gandhi, Rikin and Nambi, Akshay},
title = {Farmer.Chat: Scaling AI-Powered Agricultural Services for Smallholder Farmers},
year = {2024},
month = {October},
journal={arXiv preprint arXiv:2409.08916},
url = {https://www.microsoft.com/en-us/research/publication/farmer-chat-scaling-ai-powered-agricultural-services-for-smallholder-farmers/},
}

@inproceedings{papineni2002bleu,
author = {Papineni, Kishore and Roukos, Salim and Ward, Todd and Zhu, Wei-Jing},
title = {BLEU: a method for automatic evaluation of machine translation},
year = {2002},
publisher = {Association for Computational Linguistics},
address = {USA},
url = {https://doi.org/10.3115/1073083.1073135},
doi = {10.3115/1073083.1073135},
booktitle = {Proceedings of the 40th Annual Meeting on Association for Computational Linguistics},
pages = {311–318},
numpages = {8},
location = {Philadelphia, Pennsylvania},
series = {ACL '02}
}

@inproceedings{lin-2004-rouge,
    title = "{ROUGE}: A Package for Automatic Evaluation of Summaries",
    author = "Lin, Chin-Yew",
    booktitle = "Text Summarization Branches Out",
    month = jul,
    year = "2004",
    address = "Barcelona, Spain",
    publisher = "Association for Computational Linguistics",
    url = "https://aclanthology.org/W04-1013/",
    pages = "74--81"
}

@misc{yang2024gpt4,
  title        = {GPT-4 as Evaluator: Evaluating Large Language Models on Pest Management in Agriculture},
  author       = {Yang, Shanglong and Yuan, Zhipeng and Li, Shunbao and Peng, Ruoling and Liu, Kang and Yang, Po},
  howpublished = {arXiv preprint (arXiv:2403.11858)},
  year         = {2024},
  month        = {Mar},
  note         = {Version 1, Mar 18, 2024},
}

@article{Johnny2022FarmerQuery,
  author       = {Johnny, Swapna and Jaya Nirmala, S.},
  title        = {Farmer Query Answering System},
  journal      = {SN Computer Science},
  volume       = {3},
  number       = {1},
  pages        = {45},
  year         = {2022},
  doi          = {10.1007/s42979-021-00908-x},
}

@article{vallat2018pingouin, doi = {10.21105/joss.01026}, url = {https://doi.org/10.21105/joss.01026}, year = {2018}, publisher = {The Open Journal}, volume = {3}, number = {31}, pages = {1026}, author = {Vallat, Raphael}, title = {Pingouin: statistics in Python}, journal = {Journal of Open Source Software} }

@article{seabold2010statsmodels,
  author = {Seabold, Skipper and Perktold, Josef},
  title = {Statsmodels: Econometric and Statistical Modeling with Python},
  journal = {SciPy 2010},
  year = {2010},
  doi = {10.25080/Majora-92bf1922-011},
  url = {https://doi.org/10.25080/Majora-92bf1922-011}
}

@article{mckinney2010data,
  author = {McKinney, Wes},
  title = {Data Structures for Statistical Computing in Python},
  journal = {SciPy 2010},
  year = {2010},
  doi = {10.25080/Majora-92bf1922-00a},
  url = {https://doi.org/10.25080/Majora-92bf1922-00a}
}

@article{harris2020array,
  title={Array programming with NumPy},
  author={Harris, Charles R and Millman, K Jarrod and Van Der Walt, St{\'e}fan J and Gommers, Ralf and Virtanen, Pauli and Cournapeau, David and Wieser, Eric and Taylor, Julian and Berg, Sebastian and Smith, Nathaniel J and others},
  journal={nature},
  volume={585},
  number={7825},
  pages={357--362},
  year={2020},
  doi={https://doi.org/10.1038/s41586-020-2649-2},
  publisher={Nature Publishing Group UK London}
}

@inproceedings{zhang2019bertscore,
  title={BERTScore: Evaluating Text Generation with BERT},
  author={Tianyi Zhang* and Varsha Kishore* and Felix Wu* and Kilian Q. Weinberger and Yoav Artzi},
  booktitle={International Conference on Learning Representations},
  year={2020},
  url={https://openreview.net/forum?id=SkeHuCVFDr}
}

@inproceedings{post2018call,
    title = "A Call for Clarity in Reporting {BLEU} Scores",
    author = "Post, Matt",
    editor = "Bojar, Ond{\v{r}}ej  and
      Chatterjee, Rajen  and
      Federmann, Christian  and
      Fishel, Mark  and
      Graham, Yvette  and
      Haddow, Barry  and
      Huck, Matthias  and
      Yepes, Antonio Jimeno  and
      Koehn, Philipp  and
      Monz, Christof  and
      Negri, Matteo  and
      N{\'e}v{\'e}ol, Aur{\'e}lie  and
      Neves, Mariana  and
      Post, Matt  and
      Specia, Lucia  and
      Turchi, Marco  and
      Verspoor, Karin",
    booktitle = "Proceedings of the Third Conference on Machine Translation: Research Papers",
    month = oct,
    year = "2018",
    address = "Brussels, Belgium",
    publisher = "Association for Computational Linguistics",
    url = "https://aclanthology.org/W18-6319/",
    doi = "10.18653/v1/W18-6319",
    pages = "186--191",
}

\clearpage
\appendix

\section{Data Source}
\label{sec:data_source}
\begin{table}[H]
\centering
\small
\setlength{\tabcolsep}{8pt}
\renewcommand{\arraystretch}{1.1}

\begin{tabular}{@{}lrr@{}}
\toprule
\textbf{Topic} & \textbf{Files} & \textbf{Tokens} \\
\midrule
Agri Exports & 9 & 24,813 \\
Agri Inputs & 60 & 105,348 \\
Agri Insurance & 9 & 36,643 \\
Agro enterprises & 31 & 101,939 \\
Best Practices & 75 & 174,997 \\
Crop Production & 637 & 2,250,709 \\
Fisheries & 112 & 544,672 \\
Forestry & 18 & 69,916 \\
ICT applications\_in\_Agriculture & 18 & 36,664 \\
Livestock & 107 & 415,639 \\
Market information & 13 & 32,483 \\
National Schemes\_for\_Farmers & 17 & 72,449 \\
Policies and Schemes & 86 & 213,880 \\
Post Harvest Technologies & 41 & 158,399 \\
Poultry & 25 & 75,997 \\
State specific schemes for farmers & 50 & 117,642 \\
Women and agriculture & 42 & 87,372 \\
\midrule
\textbf{Total } & \textbf{1350} & \textbf{4,519,562} \\
\bottomrule
\end{tabular}
\caption{Token and topic statistics across different topics in the Vikaspedia dataset.}
\label{tab:token-distribution}
\end{table}

\section{Synthetic QA pair comparison between Llama and DeepSeek}
\label{appendix:dSynthetic QA pair comparison between Llama and DeepSeek}

\begin{minipage}[t]{0.43\textwidth}
\centering
\footnotesize
\begin{tabular}{p{0.13\textwidth} p{0.80\textwidth}}
\toprule
\textbf{Model} & \textbf{Question and Answer} \\
\midrule
Llama & 
\textbf{Q:} How can I manage thrips in groundnut?\\
& \textbf{A:} Apply Fipronil 2 ml/litre. \\
\addlinespace[0.5ex]
DeepSeek & 
\textbf{Q:} What should I do if my groundnut crop shows yellowish-green patches and brown necrotic areas?\\
& \textbf{A:} This could be a sign of thrips infestation. Apply Fipronil 2 ml/litre to control the infestation. \\
\bottomrule
\end{tabular}
\captionof{table}{Comparison of a single QA pair generated by Llama 3.1 and DeepSeek models.}
\label{tab:qa_comparison}
\end{minipage}

\section{Prompt for synthetic data generation}
\label{appendix:prompt_synthethic_box}

The prompt used for generation of synthetic data question-answer pairs.

\begin{figure}[H]
\centering
\scriptsize 
\begin{tcolorbox}[width=\columnwidth, colback=gray!5, colframe=black]
\ttfamily
\noindent
\textbf{Instruction:} \\
Suppose you are a knowledgeable expert in agriculture. \\

Please generate as many unique sets of corresponding question and answer pairs strictly based on the provided Context. \\
No additional context is required. \\[1ex]
\textbf{Context:} \\
\{context\} \\[1ex]
\textbf{Response:} \\
Question: [Question here] \\
Answer: [Answer here] \\[1ex]
\textbf{Examples:} \\
\{examples\}
\end{tcolorbox}
\caption{Prompt used for synthetic data generation. During prompting, the \texttt{{context}} tag is replaced with a fixed input chunk from the {agricultural dataset}.}
\end{figure}

\section{Evaluation Dataset Creation}
\label{appendix:dataCreators}
The evaluation dataset used in this study for English, Hindi and Punjabi languages was curated by Kashish Mahendra, Vidhi and Sudhir Rajput. It was manually extracted and verified from the websites given in the Table \ref{tab:Llama3_hyperparams}

\begin{table}[h!]
\footnotesize
\centering
\begin{tabularx}{\linewidth}{@{}X@{}}
\toprule
\textbf{Source and Weblink} \\
\midrule
Tamil Nadu Agricultural University: \url{http://www.agritech.tnau.ac.in/expert_system/} \\[1ex]
Union Bank of India: \url{https://www.unionbankofindia.co.in/pdf/faq-agri-products.pdf} \\[1ex]
ICAR-NRRI, Cuttack: \url{https://icar-nrri.in/faq/} \\[1ex]
ICAR-NRCP, Solapur: \url{https://nrcpomegranate.icar.gov.in/FAQ} \\[1ex]
DD Kisan YouTube Channel: \url{https://www.youtube.com/watch?v=MYkBswpxRcA} \\
\bottomrule
\end{tabularx}
\caption{Sources and their corresponding web links used for evaluation dataset creation.}
\label{tab:data_sources}
\end{table}

\section{Hindi and Punjabi QA generation methodology}
\label{appendix:transcription-methodology}
{
A Hindi test dataset of 1219 QA pairs was developed by transcribing 50 videos from the DD Kisan youTube channel\footnote{\url{https://www.youtube.com/watch?v=MYkBswpxRcA}}. The transcription was performed in a two step process. First, the audio from the publicly available videos was sourced automatically using Youtube DATA API and then transcribed using Assembly AI API. Subsequently, each transcription was manually corrected by a native Hindi speaker. GenAI 2.0 Flash ~\cite{google2025gemini2flash} was employed to generate QA pairs from 500 token segments of the transcript, resulting in 1,219 Hindi QA pairs. Subsequently, these were translated into Punjabi using IndicTrans2~\cite{gala2023indictrans2}. }

\section{Translation Quality Assessment}
\label{appendix : Translation and Transcription Quality for Hindi and Punjabi}

{The quality of the translation was carefully validated. Following the Direct Assessment methodology described in~\cite{bojar2016findings}, we evaluated both translation directions, English → Hindi and Hindi → Punjabi on adequacy using a sample of 100 sentences per direction on a 0 –100 scale. Raw mean adequacy scores were 87.26 (En→Hi) and 89.51 (Hi→Pa). The inter-rater standard deviation was 5.72, indicating relatively consistent scoring. Z-scores were computed for standardization, with mean z-scores approximating zero across both directions $\approx 10^{-16}$. These results confirm that the translated dataset is of high and reliable quality.}

\section{Translation Error Analysis and Quality Control}
\label{appendix : Translation Error Analysis and Quality Control}
{
There were several issues that we encountered during the translation process, especially Hindi to Punjabi. \textbf{Some words were mistranslated due to lexical similarity.} There were few instances where the numbers were misrepresented or altered during translation.}

 {\textbf{Repetition:} There were cases where some part of the context was translated correctly but at a certain point one word or phrase repeated continuously until the end of the translation or was truncated prematurely.}

 {\textbf{Improper Translation of Proper Nouns:} Names of companies or brands were translated literally, when they should have remained unchanged.}

 {\textbf{Punctuation Errors:} The sentence was entirely translated correctly but ended with repeated periods.}

 {However, to ensure the quality of the translated dataset, all the observed issues were addressed manually by native Hindi and Punjabi speakers. Sentences with severe errors or poor quality were removed to prevent introducing noise into the dataset. A quantitative breakdown of these errors is left for future work.}

\section{LLM Details}
\label{appendix:experimentation_details}

 {Large language models often fail to generalize uniformly across languages, particularly when trained primarily on high-resource languages like English \cite{ahuja2023mega, rohera2024indicquest, ALONSO2024102938}. \cite{ahuja2023mega} empirically demonstrate that model performance on low-resource languages is often substantially lower than on high-resource languages. This indicates that the same model cannot reliably perform equally across English and non-English languages, justifying the selection of different models optimized for each language to ensure robust performance in both high and low resource settings.}

 {During the initial experimentation phase, we evaluated a total of seven models: DeepSeek 8B, Navarasa 7B, Airavata 7B, Llama 3-8B, Llama 3.1-8B, Mistral 7B, and Gemma 7B. Due to resource constraints to perform human evaluations. We shortlisted 4 models [Airavata 7B, Llama 3 B, Deepseek 8B, Mistral 7B] among them based on our subjective observations of quality of responses such as repetition, irrelevant responses. Following the initial model selection, Llama 3.1-8B was additionally evaluated at the best-performing temperature setting ($T=0.2$) to enable a direct comparison with the best-performing model, Llama 3. Based on this evaluation, Llama 3.1-8B was subsequently included in the main manuscript, and its results are reported for completeness.}

 {Similar approach was followed for Hindi, out of the 5 models selected for English, we picked Llama 3 8B and Mistral 7B because they were the best performing models for English and included 3 more models i.e. \textbf{Llama 3.1 8B, Navarasa 7B and Gemma 8B }because of their coverage of Indic languages.}

 {For Punjabi, we selected the top 3 best performing Hindi models, again due to limited human evaluations that we could perform. Please note that our decisions were motivated to optimise our resources and performance of models.
For completeness, evaluating the same models across all languages would be desirable; however, this was beyond our available computational resources.}

\begin{table}[H]
\centering
\scriptsize
\begin{tabular}{>{\raggedright\arraybackslash}p{3.5cm} >{\centering\arraybackslash}p{1.2cm} >{\raggedright\arraybackslash}p{2.7cm}}
\toprule
\textbf{LLM Name} & \textbf{Size} & \textbf{Section(s) Used} \\
\midrule
\makecell[l]{Llama-3-8B-Instruct} & 8.03B & \makecell[l]{LLM Selection (EN, HI, PA)\\ Fine-tuning (EN)\\ Prompt Selection} \\
\makecell[l]{Mistral-7B-Instruct-v0.3} & 7.25B & \makecell[l]{LLM Selection (EN, HI)} \\
\makecell[l]{Llama-3.1-8B-Instruct} & 8.03B & \makecell[l]{LLM Selection (HI, PA)\\ Fine-tuning (HI, PA)} \\
\makecell[l]{Indic-gemma-7b-finetuned-sft-Navarasa} & 7B & \makecell[l]{LLM Selection (HI, PA)} \\
\makecell[l]{DeepSeek-R1-Distill-Llama-8B} & 8.03B & \makecell[l]{LLM Selection (EN)} \\
\makecell[l]{Airavata-7B-Instruct} & 6.87B & \makecell[l]{LLM Selection (EN)} \\
\makecell[l]{gemma-7b-it} & 7B & \makecell[l]{LLM Selection (HI)} \\
\makecell[l]{DeepSeek-R1-Distill-Llama-70B} & 70.6B & \makecell[l]{Synthetic QA Generation\\ LLM-as-a-Judge} \\
\makecell[l]{Llama-4-Scout-17B-16E-Instruct} & 109B & \makecell[l]{LLM-as-a-Judge} \\
\makecell[l]{Llama-4-Maverick-17B-128E-Instruct} & 402B & \makecell[l]{LLM-as-a-Judge} \\
\bottomrule
\end{tabular}
\caption{List of LLMs used in the study, their parameter sizes, and the sections they were used in. HI: Hindi, PA: Punjabi, EN: English.}
\label{tab:llms_used_clean}
\end{table}

\section{Prompt Selection}  
\label{appendix:prompts}  

This section provides the six tested prompts used for evaluating Mistral-7B in the Prompt Selection experiment and the design intuition behind them. Each prompt was designed to optimize the quality of the response for agricultural question-answer generation.
\\

\noindent\textbf{Prompts Tested}

\begin{itemize}
    \item \textbf{Prompt 1 (Selected Prompt)}  
    
        \textit{As an experienced agronomist proficient in farming techniques, crop management, and disease-resistant crop cultivation, you are tasked with answering questions based on your expertise. Questions may be provided in various languages such as English, Hindi, Marathi, etc.}
        
        \textit{Provide only the answer relevant to the question. Do not include any other information. The answer should be in the same language as the question.}

    \item \textbf{Prompt 2}  
    
        \textit{You are an expert in agricultural science with over 25 years of experience in farming, crop management, and sustainable agricultural practices. Your expertise spans across various domains, including soil health, irrigation techniques, pest control, and the integration of modern technology in agriculture. Your task is to answer questions related to the agriculture domain in a clear, concise, and informative manner. Ensure your responses are well-researched, accurate, and tailored to the specific context of the question. Questions may be provided in various languages such as English, Hindi and Marathi. Provide only an answer relevant to the question.}

    \item \textbf{Prompt 3}  
  
        \textit{You are an expert in the field of agriculture with decades of experience in farming, crop management, soil science, and agricultural technology. Your knowledge spans traditional and modern farming practices, and you are fluent in multiple languages, including English, Hindi, Marathi, and others. Your expertise allows you to provide accurate, practical, and culturally relevant advice tailored to the specific needs of farmers, researchers, and agricultural enthusiasts.Your task is to provide detailed, accurate, and actionable answers exclusively related to the agriculture domain. Your responses must be in the same language as the question, whether it is in English, Hindi, Marathi, or any other language. Ensure your answers are clear, concise, and directly address the query while incorporating relevant agricultural principles, practices, and innovations.}

    \item \textbf{Prompt 4}  
    
        \textit{You are a seasoned agricultural expert with over 25 years of experience in farming, crop management, and sustainable agricultural practices. Your deep understanding of soil health, pest control, irrigation systems, and modern farming technologies has made you a trusted advisor to farmers, policymakers, and agricultural organizations worldwide. Your task is to answer a specific question related to the agriculture domain. The question will be provided to you, and your response should be detailed, accurate, and tailored to the context of the query. Ensure your answer is backed by scientific evidence and a clear understanding of the latest advancements in agriculture.}

    \item \textbf{Prompt 5}  
  
        \textit{You are an expert in agriculture and tasked with answering questions related to agriculture.}\\
        \textit{Please answer the following question with detailed, accurate, and insightful information based on your knowledge in the agricultural domain.}

    \item \textbf{Prompt 6}  
   
        \textit{You are an expert agricultural advisor with deep knowledge of farming, soil health, pest control, and modern techniques. Provide a clear and well-explained response to the following question, ensuring accuracy, practicality, and ease of understanding. If multiple solutions exist, explain the best approach considering efficiency, cost-effectiveness, and sustainability. Factor in seasonal changes, soil conditions, and climate impact when relevant.}
  
\end{itemize}
\noindent\textbf{Prompt Intuition}

\begin{itemize}
    \item \textbf{Prompt 1 (Selected):} This prompt was Designed to deliver concise and contextually relevant answers in the same language as the input, ensuring minimal drift from the question's intent.

    \item \textbf{Prompt 2:} This prompt was designed to simulate a highly knowledgeable agricultural expert, embedding deep domain-specific authority into the LLM’s persona. The aim was to guide the LLM to generate precise, well-informed, and context-aware answers like domain experts.

    \item \textbf{Prompt 3:} This prompt was designed to extract detailed, accurate, and context-sensitive responses by positioning the LLM as a multilingual agricultural expert. Emphasis was placed on reasoning, cultural relevance, and linguistic alignment to ensure comprehensive, well-researched answers tailored to diverse user needs.

    \item \textbf{Prompt 4:} This prompt was designed to position the LLM as an advisor trusted by farmers and policymakers alike, having scientific expertise with leadership, so that it can provide context-rich, evidence-based answers rooted in both practical farming knowledge and awareness of modern agricultural advancements.

    \item \textbf{Prompt 5:} This prompt was kept minimal to evaluate the LLM's inherent capabilities without heavy role conditioning, by providing only a brief instruction focused on expected output quality.

    \item \textbf{Prompt 6:} The intent of this prompt was to encourage comparative reasoning and decision-making within itself. Instruct it to consider multiple solutions and select the best one based on accuracy and practicality. Ensuring responses are optimized for real-world agricultural conditions, including seasonal, soil, and climate factors.
\end{itemize}

\section{Hyperparameters and Implementation Details}
\label{appendix:hyperparameters}

\begin{table}[H]
\centering
\small
\caption{Fine-tuning hyperparameters for \textbf{Llama-3-8B-Instruct} using QLoRA.}
\begin{tabular}{@{}ll@{}}
\toprule
\textbf{Parameter} & \textbf{Value} \\
\midrule
Base LLM & Llama-3-8B-Instruct \\
LoRA Rank ($r$) & 16 \\
LoRA Alpha & 16 \\
Target Modules & \makecell[l]{q\_proj, k\_proj, v\_proj, o\_proj,\\ gate\_proj, up\_proj, down\_proj} \\
Batch Size & 2 \\
Gradient Accumulation Steps & 4 \\
Max Sequence Length & 2048 \\
Learning Rate & 2e-4 \\
Epochs & 1 \\
Optimizer & AdamW (8-bit) \\
Quantization & 4-bit (bnb) \\
Checkpointing & Enabled (Unsloth) \\
Scheduler & Linear with warm-up \\
Tracking & Weights \& Biases \\
Tokenizer & Llama 3 Instruct tokenizer \\
\bottomrule
\end{tabular}
\label{tab:Llama3_hyperparams}
\end{table}


\begin{table}[H]
\centering
\small
\caption{Fine-tuning hyperparameters for \textbf{Llama-3.1-8B-Instruct} using QLoRA.}
\begin{tabular}{@{}ll@{}}
\toprule
\textbf{Parameter} & \textbf{Value} \\
\midrule
Base LLM & Llama-3.1-8B-Instruct \\
LoRA Rank ($r$) & 16 \\
LoRA Alpha & 16 \\
Target Modules & \makecell[l]{q\_proj, k\_proj, v\_proj, o\_proj,\\ gate\_proj, up\_proj, down\_proj} \\
Batch Size & 2 \\
Gradient Accumulation Steps & 4 \\
Max Sequence Length & 2048 \\
Learning Rate & 2e-4 \\
Epochs & 1 \\
Optimizer & AdamW (8-bit) \\
Quantization & Not Used \\
Checkpointing & Enabled (Unsloth) \\
Scheduler & Linear with warm-up \\
Tracking & Weights \& Biases \\
Tokenizer & Llama 3 Instruct tokenizer \\
\bottomrule
\end{tabular}
\label{tab:Llama3_1_hyperparams}
\end{table}

We employed Parameter-Efficient Fine-Tuning (PEFT) using the QLoRA method~\cite{dettmers2023qlora} for adapting both the Llama-3-8B-Instruct and Llama-3.1-8B-Instruct LLMs. PEFT was implemented via the Unsloth framework, which enables memory-efficient training through 4-bit quantization (used only for Llama-3) and Low-Rank Adaptation (LoRA).
Gradient checkpointing was enabled to reduce memory usage during backpropagation.

Fine-tuning followed an Instruct-style prompt format and was performed using Hugging Face’s \texttt{SFTTrainer} utility.

The Inter-rater agreement was evaluated using ICC and Fleiss’ Kappa. ICC(2,k) and ICC(3,k) were computed using \texttt{Pingouin} \cite{vallat2018pingouin}, while additional ANOVA based manual ICC derivations were implemented using \texttt{statsmodels} \cite{seabold2010statsmodels} to validate agreement and consistency formulations from mean-square estimates (MSR, MSC, and MSE). Fleiss’ Kappa was computed using \texttt{statsmodels}. Data preprocessing and aggregation were performed using \texttt{Pandas} \cite{mckinney2010data} and \texttt{NumPy} \cite{harris2020array}. The manually derived ICC(3,k) consistency estimates matched the Pingouin based results, confirming the correctness and reproducibility of the inter-rater agreement computations. BERTScore, ROUGE-L, and BLEU scores were computed using the \textit{bert-score}~\cite{zhang2019bertscore}, \textit{rouge-score} ~\cite{lin-2004-rouge}, and \textit{nltk/sacrebleu}~\cite{papineni2002bleu,post2018call} Python packages, respectively.

\section{Human Evaluation Setup}  
\label{appendix:HME} 

\noindent
 {This appendix details the evaluation framework used to assess the quality of LLM-generated answers in the agricultural domain. Evaluators are native speakers of Hindi or Punjabi and are also fluent in English. The human evaluation was conducted by annotators without formal agricultural training; however, they participated in multiple piloting and training rounds to familiarize themselves with the evaluation criteria and task setup. Following this process, they were able to perform the evaluation reliably. The annotators were provided with detailed human evaluation framework guidelines and had access to the original source documents from which the question–answer pairs were derived, allowing them to verify model outputs accurately during the evaluation. The evaluation is conducted following three criteria:}

\begin{itemize}
    \item \textbf{Agricultural Consensus:} Does the response correlate with commonly accepted agricultural knowledge?
    \item \textbf{Relevancy:} Does the response demonstrate evidence of correct recall of knowledge (e.g., correct facts)?
    \item \textbf{Factuality:} Is the generated response factually accurate, partially correct, or incorrect?
\end{itemize}

\vspace{1em}
\noindent\textbf{1. Agricultural Consensus}\\
Responses are labeled as \textit{Yes} or \textit{No} depending on whether they are related to the agricultural domain.

\begin{itemize}
    \item \textbf{Yes:} Indicates the answer is clearly associated with agriculture or farming practices, regardless of its correctness.
    \item \textbf{No:} Indicates the answer is irrelevant to agriculture.
\end{itemize}

\textbf{Examples:}
\begin{itemize}
    \item \textbf{Question:} How to improve soil condition?\\
    \textbf{Predicted Answer:} There are several ways to improve soil condition such as use of organic matter, crop rotation, and maintaining pH of soil.\\
    \textbf{Agricultural Consensus:} Yes

    \item \textbf{Question:} What are some recommended places to purchase apples from?\\
    \textbf{Predicted Answer:} The Apple Store is the best place. They are very knowledgeable, sell and service all their devices, sell AppleCare and you can be hands-on with the computer before you buy it.\\
    \textbf{Agricultural Consensus:} No
\end{itemize}

\vspace{1em}
\noindent\textbf{2. Relevancy}\\
Responses are rated on a scale of 1–5 based on their relevance to the question. A score of 5 means highly relevant and accurate; a score of 1 means minimal understanding or relevance.

\textbf{Examples:}
\begin{itemize}
    \item \textbf{Question:} How to improve soil condition?\\
    \textbf{Predicted Answer:} There are several ways to improve soil condition such as use of organic matter, crop rotation, and maintaining pH of soil.\\
    \textbf{Relevancy Score:} 5

    \item \textbf{Question:} What are some recommended places to purchase apples from?\\
    \textbf{Predicted Answer:} The State of Jammu and Kashmir and Himachal Pradesh are the leading apple producing States in India.\\
    \textbf{Relevancy Score:} 2
\end{itemize}

\vspace{1em}
\noindent\textbf{3. Factuality}\\
Responses are compared with reference answers and categorized as:
\begin{itemize}
    \item \textbf{Correct:} Closely matches the reference answer.
    \item \textbf{Partially Correct:} Contains some correct information but has inaccuracies.
    \item \textbf{Incorrect:} Includes major inaccuracies or irrelevant information.
\end{itemize}

\textbf{Examples:}
\begin{itemize}
    \item \textbf{Question:} How to improve soil condition?\\
    \textbf{Reference Answer:} Techniques include use of organic matter, mulching, avoiding over-tillage, crop rotation, using cover crops, balancing soil pH, and using appropriate fertilizers.\\
    \textbf{Predicted Answer:} There are several ways to improve soil condition such as use of organic matter, crop rotation, and maintaining pH of soil.\\
    \textbf{Factuality:} Correct

    \item \textbf{Question:} Which season is suitable for apple plantation in Himachal Pradesh?\\
    \textbf{Reference Answer:} February to early May\\
    \textbf{Predicted Answer:} The best time to plant apple trees in Himachal Pradesh is between January and March.\\
    \textbf{Factuality:} Partially Correct

    \item \textbf{Question:} What is the full form of GMO?\\
    \textbf{Reference Answer:} Genetically Modified Organisms\\
    \textbf{Predicted Answer:} General Motors Operation\\
    \textbf{Factuality:} Incorrect
\end{itemize}

\section{Inter-annotator Agreement Calculation}
\label{sec:iaa}
Since the same fixed set of evaluators assessed a specific, non-random sample set for each language, the two-way mixed effects, absolute agreement, multiple raters/measurements model was used to calculate agreement scores ~\cite{KOO2016155}. According to standard interpretation, ICC values below 0.5 indicate \textit{poor reliability}, values between 0.5 and 0.75 suggest \textit{moderate reliability}, values between 0.75 and 0.9 denote \textit{good reliability}, and values above 0.9 reflect \textit{excellent reliability}. According to the interpretation scale proposed by \cite{landis-koch-1977-measurement} , the strength of agreement for Kappa statistic can be categorized with values less than 0 indicate \textit{Poor} agreement, 0--0.20 as \textit{Slight}, 0.21 -- 0.40 as \textit{Fair}, 0.41 -- 0.60 as \textit{Moderate}, 0.61 -- 0.80 as \textit{Substantial}, and 0.81 -- 1.00 as \textit{Almost Perfect}. For binary judgments related to agricultural consensus, Fleiss Kappa ~\cite{fleiss1971kappa} was used to evaluate the reliability between evaluators.

\begin{equation}
\text{ICC} = 
\frac{
  MS_R - MS_E
}{
  MS_R + \frac{MS_C - MS_E}{n}
}
\end{equation}
\vspace{1mm}\\
\noindent $MS_R$ = row mean square (subjects), 
$MS_C$ = column mean square (raters), 
$MS_E$ = error mean square, 
$n$ = number of raters per subject.

\section{Prompt Response generation }  
\label{appendix:prompt_qa_box} 

\begin{figure}[H]
\centering
\scriptsize 
\begin{tcolorbox}[width=\columnwidth, colback=gray!5, colframe=black]
\ttfamily
\noindent
\textbf{Instruction:}\newline
As an experienced agronomist proficient in farming techniques, crop management, and disease-resistant crop cultivation,
you are tasked with answering questions based on your expertise. Questions may be provided in various languages such as English, Hindi, Marathi, etc.\\
Provide only the answer relevant to the question. Do not include any other information. Answer should be in the same language as the question.

\vspace{1em}
\noindent
\textbf{Input:}\newline
Question: \{input\}

\vspace{1em}
\noindent
\textbf{Response:}
\end{tcolorbox}
\caption{Prompt used for answer generation for a given question.}
\label{fig:prompt}
\end{figure}

\section{Temperature and Prompt Selection}
\label{appendix:ModelTempSelection}

\begin{table}[H]
\centering
\small
\resizebox{\columnwidth}{!}{%
\begin{tabular}{lccc}
\toprule
\textbf{Prompt Number} & \textbf{Consensus} & \textbf{Relevancy} & \textbf{Factuality} \\
\midrule
\textbf{Prompt 1} & \textbf{100.0\%} & \textbf{95.0\%} & \textbf{55.0\%} \\
Prompt 2 & 100.0\% & 80.0\% & 50.0\% \\
Prompt 3 & 100.0\% & 85.0\% & 45.0\% \\
Prompt 4 & 100.0\% & 85.0\% & 50.0\% \\
Prompt 5 & 100.0\% & 100.0\% & 40.0\% \\
Prompt 6 & 100.0\% & 75.0\% & 40.0\% \\
\bottomrule
\end{tabular}
}
\caption{Results for the prompts for English.}
\label{tab:human_eval_results_prompt}
\end{table}

\begin{table}[H]
\small
\centering
\scriptsize
\renewcommand{\arraystretch}{1.25}
\setlength{\tabcolsep}{2.5pt} 

\begin{tabular}{
l
cc
cc
cc
}
\toprule
\multirow{2}{*}{\textbf{Model}} &
\multicolumn{2}{c}{\textbf{Agr. Consensus}} &
\multicolumn{2}{c}{\textbf{Relevance}} &
\multicolumn{2}{c}{\textbf{Factuality}} \\
& \textbf{$T=0.2$} & \textbf{$T=0$}
& \textbf{$T=0.2$} & \textbf{$T=0$}
& \textbf{$T=0.2$} & \textbf{$T=0$} \\
\midrule
Airavata              & 100.0 & 90.0  & 58.3 & 43.3 & 30.0 & 20.0 \\
Mistral 7B v0.3       & 98.3  & 95.0  & 93.3 & 75.0 & 30.0 & 40.0 \\
Llama 3 8B Instruct   & 100.0 & 95.0  & 98.3 & 76.6 & 66.6 & 43.3 \\
DeepSeek 8B Distill   & 100.0 & 95.0  & 33.3 & 66.6 & 28.9 & 40.0 \\
Llama 3.1 8B Instruct & 95.0  & --    & 58.3 & --   & 16.6 & --   \\
\bottomrule
\end{tabular}
\caption{ {Performance comparison of LLMs evaluated at temperature settings $T=0.2$ and $T=0$ across agriculture consensus, relevance, and factuality metrics.}}
\label{tab:temp_comparison}
\end{table}
\rowcolors{1}{}{}


\clearpage

\begin{table}[H]
\centering
\small
\renewcommand{\arraystretch}{1.2}
\setlength{\tabcolsep}{8pt}
\begin{tabular}{lccc|ccc|ccc}
\toprule
\textbf{LLM} 
& \multicolumn{3}{c}{\textbf{T = 0.2}} 
& \multicolumn{3}{c}{\textbf{T = 0.5}} 
& \multicolumn{3}{c}{\textbf{T = 0.7}} \\
\cmidrule(lr){2-4} \cmidrule(lr){5-7} \cmidrule(lr){8-10}
 & Con & Rel & Fact 
 & Con & Rel & Fact 
 & Con & Rel & Fact \\
\midrule
DeepSeek R1     & 100.0\% & 33.3\% & 28.3\% & 100.0\% & 80.0\% & 31.6\% & 100.0\% & 70.0\% & 21.6\% \\
\textbf{Llama 3}      & \textbf{100.0}\% & \textbf{98.3}\% & \textbf{66.6}\% & 100.0\% & 93.3\% & 56.6\% & 100.0\% & 95.0\% & 51.6\% \\
Mistral  & 100.0\% & 93.3\% & 35.0\% & 100.0\% & 75.0\% & 45.0\% & 95.0\% & 66.6\% & \ 55.0\% \\
Airavata      & 100.0\% & 58.3\% & 30.0\% & 100.0\% & 95.0\% & 41.6\% & 100.0\% & 90.0\% & 40.0\% \\
\bottomrule
\end{tabular}
\centering
\caption{English LLM evaluation across 3 different temperature (\(T\)) settings using metrics: Agriculture Consensus (Con), Factuality (Fact), and Relevancy (Rel). The highest percentage is in bold.}
\label{tab:model-eval-temp}
\end{table}

\onecolumn
\section{LLM as Judge Prompts}  
\label{appendix:llmjudge-prompts}

\begin{tcolorbox}[
    width=\textwidth,
    colback=gray!3!white,
    colframe=teal!50!black,
    coltitle=white,
    boxrule=1pt,
    arc=8pt,
    left=10pt,
    right=10pt,
    top=10pt,
    bottom=10pt,
    fonttitle=\bfseries\large,
    title=Prompt for Agriculture Consensus,
    sharp corners=south,
]
\textbf{Instruction:} \\
\small{You will be given a question, a reference (gold) answer, and an LLM-generated answer. \\
Your task is to rate the LLM-generated responses on one metric. \\
Please make sure you read and understand these instructions carefully. \\

Evaluation Criteria: \\
Agricultural Consensus (Yes,No) - To assess the results generated by the LLM, we categorize them with labels of Yes or No depending on their correlation with the agriculture sector. \\
A Yes label denotes a clear association with agricultural or farming activities, regardless of the accuracy of the response. Conversely, a No label indicates a lack of relevance to agricultural domain knowledge. \\
NOTE: The generated response may not be accurate, but it should be relevant to the agriculture field. \\[1ex]

\textbf{Evaluation Steps:} \\
1. Understand the Purpose: \\
\hspace*{1em}- Recognize that the goal is to determine whether the LLM-generated response is relevant to the agriculture or farming domain, not whether it is factually accurate. \\

2. Read the Question: \\
\hspace*{1em}- Review the question to understand the context in which the LLM was expected to respond. This can provide cues about whether the topic is related to agriculture. \\

3. Analyze the Reference Answer: \\
\hspace*{1em}- Briefly examine the reference (gold) answer to see what type of response is expected and whether it is centered around agricultural knowledge or practices. \\

4. Review the LLM-Generated Answer: \\
\hspace*{1em}- Carefully read the generated answer to determine the domain it is addressing. Identify key terms, phrases, or themes that indicate its association (or lack thereof) with agriculture or farming. \\

5. Assess Domain Relevance: \\
\hspace*{1em}- Check for direct or indirect connections to agricultural topics such as crops, soil, irrigation, farming techniques, pest control, agricultural policy, rural economy, livestock, climate effects on farming, etc. \\

6. Make a Binary Decision: \\
\hspace*{1em}- If the generated answer clearly pertains to the agriculture sector, assign the label "Yes". \\
\hspace*{1em}- If the generated answer is unrelated to agriculture, farming, or associated practices, assign the label "No". \\

7. Avoid Judging Accuracy: \\
\hspace*{1em}- Do not consider factual correctness when assigning the label. The only criterion is whether the content is related to the agriculture domain. \\[1ex]

\textbf{Inputs:} \\
Question: \{question\} \\

Reference (Gold) Answer: \\
\{answer\} \\

LLM-Generated Answer: \\
\{answer\_llm\}}
\end{tcolorbox}

\twocolumn

\begin{figure*}[t]
\begin{tcolorbox}[
    width=1.00\textwidth,
    colback=gray!3!white,
    colframe=teal!50!black,
    coltitle=white,
    boxrule=1pt,
    arc=8pt,
    left=10pt,
    right=10pt,
    top=10pt,
    bottom=10pt,
    fonttitle=\bfseries\large,
    title=Prompt for Factuality Evaluation,
    sharp corners=south,
]

\small{
\textbf{Instruction:} \\
You will be given a question, a reference (gold) answer, and an LLM-generated answer. \\
Your task is to rate the LLM-generated responses on one metric. \\
Please make sure you read and understand these instructions carefully. \\[1ex]

\textbf{Evaluation Criteria:} \\
Factuality (Correct, Partially Correct, Incorrect) — The response generated by the LLM will be assessed against the reference answer to evaluate its accuracy. \\
It will be classified as either \textbf{"Correct"}, \textbf{"Partially Correct"}, or \textbf{"Incorrect"}. \\
The \textbf{Correct} designation denotes a response that closely matches the reference answer, while an \textbf{Incorrect} label indicates significant inaccuracies or inappropriate information. \\
\textbf{Partially Correct} refers to a response that is not entirely accurate but includes some elements or pieces of information that are correct. \\[1ex]

\textbf{Evaluation Steps:} \\

1. \textbf{Understand the Question:} \\
\hspace*{1em}- Read the question carefully to ensure a proper understanding of what is being asked. Clarify any ambiguity in the question to accurately evaluate the generated response. \\

2. \textbf{Review the Reference Answer (Gold Answer):} \\
\hspace*{1em}- Thoroughly analyze the reference answer to understand the key details and factual information it presents. Identify the core points, facts, or context that must be preserved in the LLM-generated answer. \\

3. \textbf{Review the LLM-Generated Answer:} \\
\hspace*{1em}- Read the LLM-generated response closely. Check if the information presented aligns with the reference answer. Identify the factual elements in the response and compare them with the reference answer to spot any discrepancies or omissions. \\

4. \textbf{Assess Factual Accuracy:} \\
\hspace*{1em}- Compare each key detail in the LLM-generated answer with the reference answer. \\
\hspace*{1em}- If the LLM-generated answer fully matches the factual information provided in the reference answer, rate it as \textbf{Correct}. \\
\hspace*{1em}- If the LLM-generated answer has some elements that align with the reference answer but also contains inaccuracies or incomplete information, rate it as \textbf{Partially Correct}. \\
\hspace*{1em}- If the LLM-generated answer includes significant inaccuracies or irrelevant information, rate it as \textbf{Incorrect}. \\

5. \textbf{Contextualize the Answer:} \\
\hspace*{1em}- Ensure that the context of the answer (e.g., domain knowledge, specific terminology) is understood and correctly represented. If the response introduces concepts that are out of context or misleading, it should be rated as \textbf{Incorrect}. \\

6. \textbf{Check for Detail and Completeness:} \\
\hspace*{1em}- Evaluate the level of detail in the response. If critical facts or key elements are missing from the LLM's response compared to the reference answer, it should be considered \textbf{Partially Correct} or \textbf{Incorrect}, depending on the extent of the missing information. \\

7. \textbf{Final Rating:} \\
\hspace*{1em}- Assign the final rating to the LLM-generated answer based on the accuracy of the information provided compared to the reference answer. \\[1ex]

\textbf{Inputs:} \\
Question: \{question\} \\

Reference (Gold) Answer: \\
\{answer\} \\

LLM-Generated Answer: \\
\{answer\_llm\}
}
\end{tcolorbox}
\end{figure*}

\begin{figure*}[t]
\begin{tcolorbox}[
    width=1.00\textwidth,
    colback=gray!3!white,
    colframe=teal!50!black,
    coltitle=white,
    boxrule=1pt,
    arc=8pt,
    left=10pt,
    right=10pt,
    top=10pt,
    bottom=10pt,
    fonttitle=\bfseries\large,
    title=Prompt for Relevance Evaluation,
    sharp corners=south,
]

\small{
\textbf{Instruction:} \\
You will be given a question and an LLM-generated answer. \\
Your task is to rate the LLM-generated responses on one metric. \\
Please make sure you read and understand these instructions carefully. \\[1ex]

\textbf{Evaluation Criteria:} \\
\textbf{Relevance (1–5)} — The response will be rated on a scale of 1 to 5 based on its relevance to the question. \\
A rating of \textbf{5} indicates a highly relevant and well-aligned response, while a rating of \textbf{1} indicates that the answer is largely irrelevant or mismatched with the question. \\
\textbf{NOTE:} The answer must align with the question asked, regardless of its factual accuracy. A response that incorporates all pertinent information and clearly addresses the core intent of the question will score higher. \\[1ex]

\textbf{Evaluation Steps:} \\

1. \textbf{Understand the Question:} \\
\hspace*{1em}- Read the question carefully to identify the core intent and expected scope of the answer. Note keywords and concepts that define what kind of information is being asked. \\

2. \textbf{Read the LLM-Generated Answer:} \\
\hspace*{1em}- Examine the generated answer in its entirety. Determine what topic it is addressing, what information it provides, and how closely it connects to the original question. \\

3. \textbf{Match Answer to Question Intent:} \\
\hspace*{1em}- Check whether the LLM-generated answer addresses the main point of the question. \\
\hspace*{1em}- Consider if the answer is directly responsive to the question, or if it strays into unrelated or tangential topics. \\

4. \textbf{Evaluate Depth and Completeness:} \\
\hspace*{1em}- Assess whether the response includes not just a relevant answer but also supporting details, context, or additional information that enhances its relevance. \\
\hspace*{1em}- Responses that include a broader context or deeper insight into the question’s topic may score higher. \\

5. \textbf{Rate on a Scale from 1 to 5:} \\
\hspace*{1em}- \textbf{5:} Fully relevant and directly answers the question with complete, well-contextualized information. \\
\hspace*{1em}- \textbf{4:} Mostly relevant and answers the question well, with minor gaps or slight vagueness. \\
\hspace*{1em}- \textbf{3:} Moderately relevant; some parts address the question, but the answer is incomplete or only partially focused. \\
\hspace*{1em}- \textbf{2:} Minimally relevant; contains limited information related to the question or is largely off-topic. \\
\hspace*{1em}- \textbf{1:} Irrelevant or barely addresses the question; shows poor understanding or misunderstanding of the question asked. \\

6. \textbf{Ignore Factual Accuracy:} \\
\hspace*{1em}- Do not factor in correctness of facts. Focus strictly on alignment with the question’s subject and intent. \\[1ex]

\textbf{Inputs:} \\
Question: \{question\} \\

LLM-Generated Answer: \\
\{answer\_llm\}
}
\end{tcolorbox}

\end{figure*}

\clearpage

\section{Misleading Score}
\label{appendix:misleadingSCORES}

\begin{table}[H]
\centering
\scriptsize
\renewcommand{\arraystretch}{1.3}
\setlength{\tabcolsep}{4pt}
\begin{tabular}{|p{3.5cm}|p{3.5cm}|p{3.5cm}|c|c|c|c|c|}
\hline
\textbf{Question} & \textbf{Gold Answer} & \textbf{LLM Answer} & \textbf{Human} & \textbf{LLM} & \textbf{BERT} & \textbf{ROUGE-L} & \textbf{BLEU} \\
\hline

What is the cause for round scales on the under surface of leaves along with yellowing of leaves on the upper surface in coconut palm? &
Scale insects reside on the under surface of leaves and cause yellowing on the upper surface of leaves. They secrete wax in circular patches. &
Fungal infection \textit{(Colletotrichum gloeosporioides)} and insect infestation \textit{(Necrobia rufipes)} are the common causes of round scales on the under surface of leaves along with yellowing of leaves on the upper surface in coconut palm. &
Incorrect &
Partially Correct &
0.6710 &
0.4482 &
0.2216 \\
\hline

Is pseudostem split symptom caused by wilt pathogens? &
Yes, the FOC can cause pseudostem split symptoms. Similar symptoms are also noticed with Erwinia rot disease. &
No, pseudostem split symptom is not caused by wilt pathogens. It is caused by the infestation of the stem borer, \textit{Chilo partellus}. &
Incorrect &
Incorrect &
0.650 &
0.153 &
0.030 \\
\hline

Is pseudostem split symptom caused by wilt pathogens? &
Yes, the FOC can cause pseudostem split symptoms. Similar symptoms are also noticed with Erwinia rot disease. &
Yes, it is caused by wilt pathogens. &
Correct &
Correct &
0.666 &
0.166 &
0.007 \\
\hline

Can I apply the entire fertilizer recommended at one time for coconut plantation? &
Apply in split doses to ensure regular nutrient availability and avoid leaching. &
No, apply in controlled manner over time for optimal growth. &
Correct &
Partially Correct &
0.621 &
0.170 &
0.019 \\
\hline

Can I apply the entire fertilizer recommended at one time for coconut plantation? &
Apply in split doses to ensure regular nutrient availability and avoid leaching. &
Yes, you can apply the entire fertilizer recommended at one time. &
Incorrect &
Incorrect &
0.567 &
0.091 &
0.000 \\
\hline
\end{tabular}
\caption{Evaluation of LLM-generated answers against gold references with human and LLM annotations along with automatic scores.}
\label{tab:qa-eval-full}
\end{table}

\clearpage

\section{LLM-as-a-Judge : Pearson Correlation for Agriculture Consensus Metric}
\label{appendix:PearsonAgriConsensus}
For the agriculture consensus metric, which involves binary classification, Pearson correlation~\cite{pearson1895correlation, fu-etal-2023-large} was considered. However, the correlation yielded a \texttt{NaN} value due to the absence of variance in the evaluations: both the annotators and the LLM provided identical labels for all instances, indicating complete agreement.

\section{Hardware and Infrastructure}
All experiments were conducted on high-performance GPU nodes equipped with NVIDIA Tesla V100 (32GB) and NVIDIA A100 (80GB) GPUs. The environment was configured to efficiently support large-scale LLM inference and fine-tuning, leveraging mixed precision and distributed processing where applicable.

\begin{table}[H]
\centering
\scriptsize  
\begin{tabular}{@{}p{5.5cm}rr@{}}
\toprule
\textbf{Experiment Description} & \textbf{Hours} & \textbf{Minutes} \\
\midrule

\textbf{Fine-tuning} & & \\
\hspace{2mm}Punjabi Llama3.1 Finetuning & 3 & 12 \\
\hspace{2mm}Hindi Llama3.1 Finetuning & 2 & 17 \\
\hspace{2mm}English Llama3 Finetuning & 2 & 28 \\

\midrule
\textbf{Model and Temperature Selection} & & \\
\hspace{2mm}Airavata & 12 & 03 \\
\hspace{2mm}DeepSeek & 29 & 06 \\
\hspace{2mm}Llama3 & 17 & 36 \\
\hspace{2mm}Mistral & 12 & 09 \\

\midrule
\textbf{Fine-tuning Inference} & & \\
\hspace{2mm}English & 1 & 46 \\
\hspace{2mm}Hindi & 2 & 36 \\
\hspace{2mm}Punjabi & 4 & 55 \\

\midrule
\textbf{Individual Model Evaluations} & & \\
\hspace{2mm}Llama3 (HI) & 4 & 41 \\
\hspace{2mm}Llama3.1 (HI) & 7 & 54 \\
\hspace{2mm}Gemma (HI) & 1 & 59 \\
\hspace{2mm}Mistral (HI) & 6 & 11 \\
\hspace{2mm}Navrasa (HI) & 4 & 36 \\
\hspace{2mm}Punjabi (HI) & 4 & 55 \\
\hspace{2mm}Llama3 (PA) & 5 & 18 \\
\hspace{2mm}Llama3.1 (PA) & 6 & 21 \\
\hspace{2mm}Navrasa (PA) & 1 & 47 \\

\midrule
\textbf{Prompt Selection Experiments} & & \\
\hspace{2mm}Prompt 1 & 5 & 55 \\
\hspace{2mm}Prompt 2 & 6 & 02 \\
\hspace{2mm}Prompt 3 & 6 & 44 \\
\hspace{2mm}Prompt 4 & 10 & 40 \\
\hspace{2mm}Prompt 5 & 11 & 43 \\
\hspace{2mm}Prompt 6 & 8 & 04 \\

\midrule
\textbf{Indic Translation} & 5 & 48 \\

\midrule
\textbf{Total GPU Time} & \textbf{178} & \textbf{46} \\
\bottomrule
\end{tabular}
\caption{Detailed breakdown of GPU time consumed across all experiments including fine-tuning, inference, model selection, and prompt evaluation.}
\label{tab:gpu_time_summary_acl}
\end{table}

\clearpage

\end{document}